\documentclass[letterpaper]{article} 
\usepackage{aaai2026}  
\usepackage{times}  
\usepackage{helvet}  
\usepackage{courier}  
\usepackage[hyphens]{url}  
\usepackage{graphicx} 
\usepackage{natbib}  
\usepackage{caption} 
\usepackage{algorithm}
\usepackage{algorithmic}

\usepackage{amsmath}   
\usepackage{amssymb}

\usepackage{subcaption}
\usepackage{amsmath}
\usepackage{booktabs}
\usepackage{newfloat}
\usepackage{listings}
\DeclareCaptionStyle{ruled}{labelfont=normalfont,labelsep=colon,strut=off} 
\floatstyle{ruled}
\newfloat{listing}{tb}{lst}{}
\floatname{listing}{Listing}
\title{RMAdapter: Reconstruction-based Multi-Modal Adapter for\\
Vision-Language Models}
\author{
    Xiang Lin\textsuperscript{\rm 1,2}, Weixin Li\textsuperscript{\rm 1,2}\thanks{Corresponding Author. Email: weixinli@buaa.edu.cn},
    Shu Guo\textsuperscript{\rm 3},
    Lihong Wang\textsuperscript{\rm 3},
    Di Huang\textsuperscript{\rm 1,2}
}
\affiliations{
    \textsuperscript{\rm 1}SKLCCSE, Beihang University, Beijing, China\\
    \textsuperscript{\rm 2}School of Computer Science and Engineering, Beihang University, Beijing, China\\
    \textsuperscript{\rm 3}National Computer Network Emergency Response Technical Team/Coordination Center of China, Beijing, China

}

\usepackage{bibentry}

\begin{document}

\maketitle

\begin{abstract}
Pre-trained Vision-Language Models (VLMs), \textit{e.g.} CLIP, have become essential tools in multimodal transfer learning.  
However, fine-tuning VLMs in few-shot scenarios poses significant challenges in balancing task-specific adaptation and generalization in the obtained model.
Meanwhile, current researches have predominantly focused on prompt-based adaptation methods, leaving adapter-based approaches underexplored and revealing notable performance gaps.  
To address these challenges, we introduce a novel Reconstruction-based Multimodal Adapter (RMAdapter), which leverages a dual-branch architecture.  
Unlike conventional single-branch adapters, RMAdapter consists of:  
(1) an adaptation branch that injects task-specific knowledge through parameter-efficient fine-tuning, and  
(2) a reconstruction branch that preserves general knowledge by reconstructing latent space features back into the original feature space.  
This design facilitates a dynamic balance between general and task-specific knowledge.
Importantly, although RMAdapter introduces an additional reconstruction branch, it is carefully optimized to remain lightweight. By computing reconstruction loss locally at each layer and sharing projection modules, the overall computational overhead is kept minimal.
A consistency constraint is also incorporated to better regulate the trade-off between discriminability and generalization.
We comprehensively evaluate the effectiveness of RMAdapter on three representative tasks: generalization to new categories, generalization to new target datasets, and domain generalization.  
Without relying on data augmentation or duplicate prompt designs, our RMAdapter consistently outperforms state-of-the-art approaches across all evaluation metrics.  
\end{abstract}

\section{Introduction}
\label{sec:intro}

\begin{figure}[t]
   \centering
   \centerline{\includegraphics[width=0.25\textwidth]{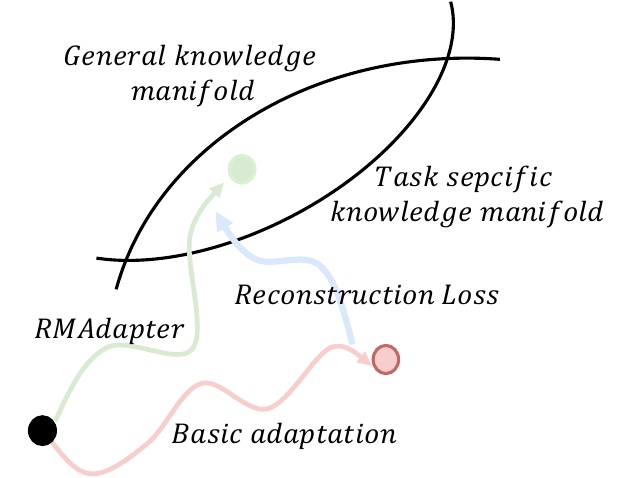}}
   \caption{Conceptual illustration of our method. Existing adaptation methods rely on task-specific optimization objectives, which leads to the loss of generalizable knowledge (pink line). Our RMAdapter (green line) preserves generalizable knowledge through the introduction of a reconstruction loss (blue line). It guides the training trajectory toward the point between two optimal solution manifolds (green dot) while learning task-specific representations.}
   \label{fig0}
\end{figure}

Vision-Language Models (VLMs) \cite{Jia2021,Yao2021Filip,Yuan2021, Zhai2022,Yu2022} have rapidly advanced in recent years, achieving promising performance in various tasks.
CLIP (Contrastive Language-Image Pretraining) \cite{radford2021learning} is a prominent VLM that learns a unified alignment space for language and visual features through large-scale cross-modal contrastive learning. By effectively bridging the semantic gap between texts and images, CLIP enables robust open-vocabulary recognition and strong generalization to unseen visual concepts \cite{Zhou2022Detecting, Gu2021}.
In its original design, CLIP employs a fixed handcrafted prompt template, \textit{e.g.} ``a photo of a \texttt{<}category\texttt{>}'', to generate text-based class embeddings for zero-shot predictions.
Although these static prompts capture general textual knowledge and support strong generalization, they lack task-specific nuances, resulting in suboptimal discrimination in downstream tasks.
Meanwhile, the model’s large-scale architecture and the limited availability of training data in few-shot scenarios make full fine-tuning for specific tasks impractical.
\begin{figure}[htbp]
\centerline{\includegraphics[width=0.47\textwidth]{rma.pdf}}
\caption{Structural evolution from the standard adapter and AutoEncoder to our proposed RMAdapter. \textbf{Left}: The internal structure of a general adapter module. The input \( x \) undergoes a down projection $W_{down}$, a nonlinear activation, and an up projection $W_{up}$ to obtain \( \text{Adapter}(x) \), which is then fused with the input \( x \) through a residual connection.
\textbf{Middle}: The general structure of an AutoEncoder. The input \( x \) is passed through an encoder \( g \) to obtain a latent representation \( \text{z} \), which is then reconstructed by a decoder \( f \) to produce \( \hat{x} \). The reconstruction loss $argmin_{f,g}L(x,\hat{x})$ is then computed to optimize the model.
\textbf{Right}: The general structure of our RMAdapter. Inspired by the structural similarity between adapters and AutoEncoders, RMAdapter employs a dual-branch architecture by integrating AutoEncoder branch to reconstruct input \( x \) to preserve general knowledge, while the \( W_{down} \) parameters are shared to further enhance the model’s performance.}
\label{fig1}
\end{figure}

To address these challenges, researchers have recently explored Parameter-Efficient Fine-Tuning (PEFT) strategies, primarily focusing on Prompt Learning \cite{Jia2022, Lester2021,Bulat2023} and Adapter-based methods \cite{Chen2022Adaptformer,Hu2021,Mou2023, Zhang2021TipAdapter, Gao2023}. 
Prompt Learning techniques \cite{Zhou2022ConditionalPrompt, Zhou2022LearningPrompt,Khattak2023,Khattak2023SelfReg,yao2024tcp,roy2024consistencyguidedpromptlearningvisionlanguage} introduce learnable prompt vectors to replace static handcrafted prompts, enabling the model to adapt dynamically to different tasks. 
Although this approach enhances task-specific discriminability, it also introduces an inherent trade-off: the knowledge acquired from few-shot samples tends to be highly discriminative for seen categories but biased against unseen ones, leading to performance degradation in unseen domains. 
This phenomenon is common in prompt-based methods, where the adaptation of task-specific knowledge often comes at the cost of forgetting general knowledge, thereby compromising the model's zero-shot generalization ability. 
On the other hand, adapter-based methods improve adaptability by integrating lightweight modules into the model \cite{Zhang2021TipAdapter, Gao2023}.  
However, most of them focus on single-modal optimization and lack explicit mechanisms for cross-modal interaction.  
Multimodal Adapters (MMA) \cite{yang2024mma} attempt to address this limitation recently by incorporating shared projection layers to enhance alignment between image and text representations.  
Nevertheless, they still struggle to effectively balance task-specific discriminability and generalization.  
Overall, despite advancements in both prompt-based and adapter-based methods, achieving an optimal trade-off between task-specific adaptation and generalization remains a significant challenge, as shown in Figure \ref{fig0}. The key issue, therefore, lies in designing adaptation strategies to effectively capture task-specific knowledge while preserving the model’s generalization capability.

To address this challenge, we propose a Reconstruction-based Multimodal Adapter (RMAdapter) in this paper.  
As illustrated in Figure \ref{fig1},
RMAdapter employs a dual-branch architecture, including (1) an Adaptation Branch, which injects task-specific knowledge through parameter-efficient fine-tuning, and (2) a Reconstruction Branch, which preserves general knowledge by imposing feature space constraints to remap latent features back to their original distribution.
By leveraging this structural innovation, RMAdapter achieves a dynamic trade-off between task-specific adaptation and general knowledge retention, ensuring both effective few-shot learning and robust zero-shot generalization. Furthermore, we reveal that sharing the down-projection layer achieves a Pareto-optimal trade-off between adaptation and reconstruction in few-shot settings. Additionally, we introduce a consistency constraint, validated in \cite{Khattak2023SelfReg,roy2024consistencyguidedpromptlearningvisionlanguage}, to further refine the trade-off between discriminability and generalization.

We comprehensively evaluate our RMAdapter on three representative tasks: Generalization from Base to-Novel Classes, Cross-dataset Evaluation, and Domain Generalization. RMAdapter consistently outperforms state-of-the-art approaches across all evaluation metrics, demonstrating its effectiveness in both few-shot and zero-shot scenarios.

In summary, our contributions are as follows:
\begin{itemize}
    \item We propose a novel dual-branch adapter architecture which dynamically balances task-specific adaptation and generalization, enabling few-shot adaptation without sacrificing zero-shot capabilities.
    \item We introduce a hierarchical sharing strategy and consistency constraints across both branches to further enhance generalization performance in few-shot settings.
    \item Experimental results demonstrate that our RMAdapter consistently surpasses prior approaches, setting new state-of-the-art performance.
\end{itemize}

\section{Related Work}
\subsection{Vision-Language Models}
In recent years, large-scale Vision-Language Models (VLMs) have exhibited remarkable potentials for multimodal understanding and reasoning \cite{radford2021learning, Jia2021, Yao2021Filip, Yuan2021}.
These models are typically pre-trained on massive paired image-text datasets collected from the internet using self-supervision, and often use a contrastive loss to pull together matched image-text features and push apart those unmatched.
For instance, CLIP \cite{radford2021learning} and ALIGN \cite{Jia2021} respectively employ about 400 million and 1 billion image-text pairs for training. By virtue of their ability to understand open-vocabulary concepts, VLMs not only perform well in few-shot or zero-shot visual recognition but have also been successfully applied to various tasks, \textit{e.g.} image classification \cite{Zhou2022Detecting}, object detection \cite{Gao2023}, segmentation \cite{Chen2022}, \textit{etc}. Nevertheless, these models are typically very large in scale. Fully fine-tuning them on downstream tasks often leads to a deterioration in their original generalization capability, while linear probing typically does not perform sufficiently well in downstream adaptation \cite{Khattak2023}. Hence, recent studies have focused on efficiently adapting foundational models without modifying their pre-trained weights. The primary solutions involve prompt learning \cite{Zhou2022ConditionalPrompt,Khattak2023,Khattak2023SelfReg,roy2024consistencyguidedpromptlearningvisionlanguage} and incorporating adapters \cite{Zhang2021TipAdapter,Gao2023,yang2024mma}.
\subsection{Prompt Learning}
Prompt Learning is originated in Natural Language Processing (NLP) \cite{Lester2021} and has later been introduced to both vision-language models \cite{Bahng2022,Zhou2022LearningPrompt,Zhou2022ConditionalPrompt} and purely vision-based models \cite{Tsimpoukelli2021,Jia2022}. Its core idea is to add a small number of learnable prompt tokens at the input or inner layers to allow the original model weights to remain frozen while providing additional tunable parameters for downstream adaptation.
CoOp \cite{Zhou2022LearningPrompt} introduces continuous prompts into the language branch for few-shot recognition but struggles with unseen classes. CoCoOp \cite{Zhou2022ConditionalPrompt} conditions prompts on image features to enhance generalization. KgCoOp \cite{yao2023visual} reduces the gap between general and specific textual embeddings for better transferability. MaPLe \cite{Khattak2023} jointly learns prompts in both modalities for multimodal synergy. PromptSRC \cite{Khattak2023SelfReg} adds self-regulation to encourage task-agnostic generalization.
Coprompt  \cite{roy2024consistencyguidedpromptlearningvisionlanguage} integrates both the prompt and adapter methods to optimize additional parameters to enhance performance and also applies a consistency constraint to potentially improve the model’s capabilities in zero-shot learning scenarios.
2SFS \cite{farina2025rethinking} is a two-stage PEFT framework that separates feature adaptation and classification, enabling efficient and effective few-shot learning.
MMRL \cite{guo2025mmrl} introduces a shared, modality-agnostic representation space and feature alignment strategy to enhance few-shot adaptation of vision-language models while preserving generalization.
Although these methods implicitly mitigate overfitting and maintain better generalization capabilities to some extent, they lack explicit structural designs to control the balance between discriminative ability and generalization performance.
\subsection{Adapter Methods}
Adapters are usually placed inside the network as a shallow feature transformation module \cite{Zhang2021TipAdapter, Chen2022Adaptformer, Gao2023, yang2024mma}.
Representative works \textit{e.g.} Clip-Adapter\cite{Gao2023}, Clip-Adapter \cite{Zhang2021TipAdapter} and AdaptFormer \cite{Chen2022Adaptformer} fuse outputs of the pre-trained model with outputs of adapters, thereby adapting the model to downstream tasks while preserving the original knowledge of the base network.
However, earlier adapter approaches primarily focus on single-modal optimization and lack explicit cross-modal interaction mechanisms. 
Recent studies have sought to combine multimodal interactions or more flexible feature alignment strategies. The Cross-Modal Adapter is introduced for text-video retrieval \cite{Jiang2022}, allowing encoder-level implicit cross-modal interactions. Multimodal Adapters (MMA) \cite{yang2024mma} aim to bridge this gap by incorporating shared projection layers, facilitating enhanced alignment between image and text representations.
Although adapters mitigate some of the overfitting issues, they still struggle to simultaneously maintain discrimination and generalization. Compared with prompt method, adapter based methods are still far from full exploration. Our work aims to address this problem by proposing a novel reconstruction adapter framework to effectively balance discrimination and generalization.

\section{Method}
Following prior researches \cite{Khattak2023SelfReg, yang2024mma,roy2024consistencyguidedpromptlearningvisionlanguage}, we utilize the pre-trained transformer-based CLIP model \cite{radford2021learning}. In the following subsections, we first provide an overview of CLIP and key concepts related to AutoEncoder. Then we delve into a detailed explanation of our proposed RMAdapter.
\begin{figure*}[t]
  \centering
   \centerline{\includegraphics[width=0.8\textwidth]{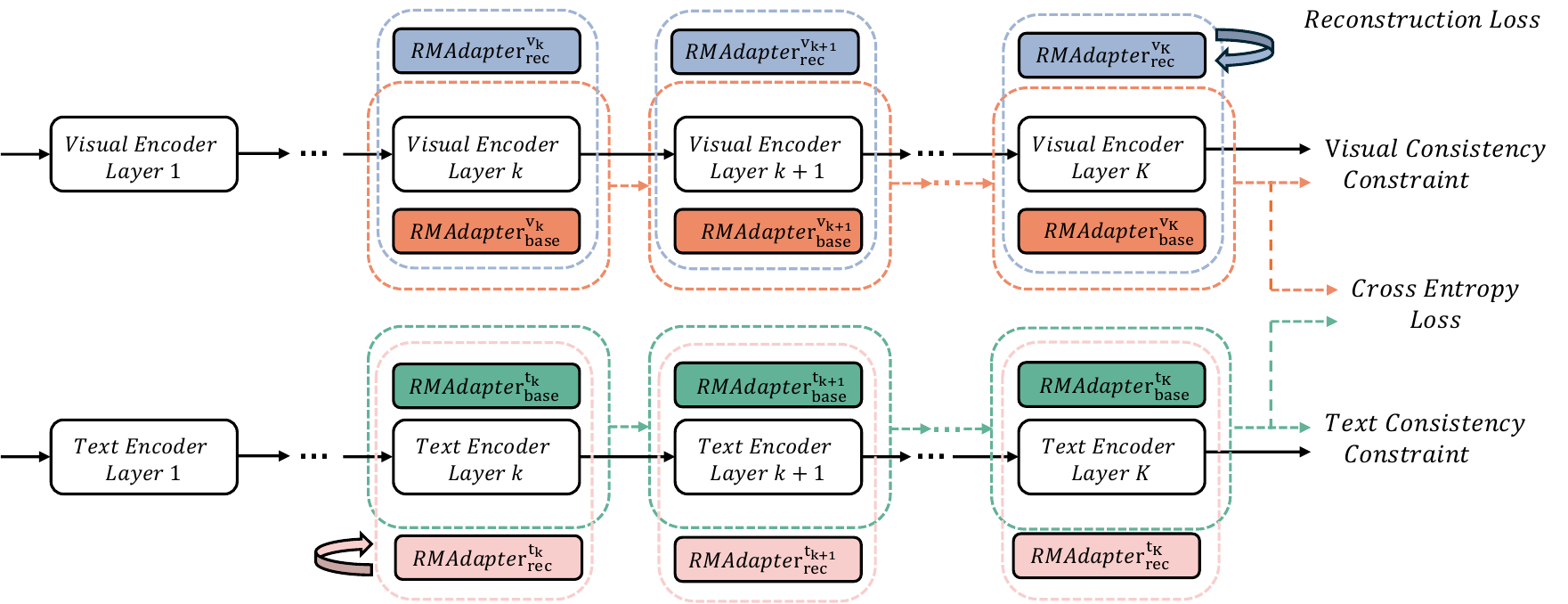}}
   \caption{The framework of our RMAdapter. RMAdapter optimizes only the additional adapters (colored parts), while the entire pre-trained CLIP model remains frozen. RMAdapter employs a dual-branch architecture consisting of: (1) an adaptation branch, $\text{RMAdapter}_{\text{base}}$ and (2) a reconstruction branch, $\text{RMAdapter}_{\text{rec}}$. Notably, the reconstruction loss is computed locally within each layer, without the need for layer-wise backpropagation or inter-layer transmission, making the computation highly efficient. Similar to previous methods, we fine-tune only the higher layers $k$ of each encoder to achieve a better balance between discriminability and generalization. The orange and green lines represent the adapted outputs, while the black lines indicate the original CLIP outputs.
   }
   \label{fig3}
\end{figure*}
\subsection{CLIP Preliminaries}
CLIP is a foundational VLM that has gained significant traction in both natural language processing and computer vision. It comprises two core components: a text encoder  and a vision encoder.
These encoders are jointly pre-trained on large-scale web-sourced image-text pairs using a contrastive learning objective \cite{Oord2018}, ensuring that semantically aligned image-text pairs are mapped closer in the embedding space while unrelated pairs are pushed apart.

\subsubsection{Image Encoder}
Given an input image $I\in \mathbb{R}^{H\times W \times 3}$, the vision encoder $\{\mathcal{V}_i\}_{i=1}^{K}$ extracts its feature representation as follow. 
The input image $I$ is first split into $M$ patches, and then these patches are projected into features $E_0 \in \mathbb{R}^{M\times h_v}$ by a $PatchEmbed$ function.
Patch embeddings $E_i$ are then input to the $(i + 1)^{th}$ transformer block $V_{i+1}$ along with a learnable class (CLS) token $c_i$ and sequentially processed through $K$ transformer blocks as:
\begin{equation}
[c_i, E_i] = \mathcal{V}_i([c_{i-1}, E_{i-1}]), \quad i = 1,2, \dots, K.
\label{eq1}
\end{equation}
To obtain the final image representation $x$, the class token $c_K$ of last transformer layer $V_K$ is projected to a common V-L latent embedding space $h_l$ via a function $ImageProj$ as
$x = ImageProj(c_K)$.

\subsubsection{Text Encoder} 
Similarly, the text encoder in CLIP $\{\mathcal{T}_i\}_{i=1}^{K}$ generates feature representations for text description by tokenizing the words and projecting them to word embeddings
$W_0 = [w_0^1, w_0^2, \dots, w_0^N]\in \mathbb{R}^{N \times h_t}$, where $N$ is the length of text.
At each stage, $W_i$ is input to the $(i + 1)^{th}$ transformer layer of text encoding branch $\mathcal{T}_{i+1}$ as: 
\begin{equation}
 W_i = \mathcal{T}_i, \quad i = 1,2, \dots, K.
  \label{eq2}
\end{equation}
The final text representation $w$ is obtained by projecting the text embeddings corresponding to the last token of the last transformer block $\mathcal{T}_K$ to a same latent embedding space $h_l$ via the $TextProj$ function as $w = TextProj(w_K^N)$.

\subsubsection{Zero-shot Classification} 
For zero-shot classification, text prompts with class $\{1,2,...,C\}$ are hand-crafted  with class labels (\textit{e.g.} ``a photo of a \texttt{<}category\texttt{>}'').
Given those features, cosine similarity scores with image representation $x$ are computed as:
\begin{equation}
p(y|x) = \frac{\exp(\operatorname{sim}(x, w_{y})/\tau)}{\sum_{k=1}^{C} \exp(\operatorname{sim}(x, w_k)/\tau)},
  \label{eq3}
\end{equation}
where $y$ corresponds to the prediction label of image $I$ and $\tau$ is a temperature parameter.

\subsection{AutoEncoder Method}
AutoEncoders (AEs) \cite{pmlr-v27-baldi12a} are a class of neural networks designed for unsupervised representation learning by reconstructing input data through an encoder-decoder framework. The encoder maps the input into a lower-dimensional latent space, while the decoder reconstructs the original input from this latent representation. The objective of an AutoEncoder is to learn a compressed and meaningful feature representation that captures essential characteristics of the input data.
Mathematically, given an input \( x \in \mathbb{R}^d \), the encoding process can be formulated as:
\begin{equation}
z = f_{\theta}(x) = \sigma(W_{\text{enc}} x + b_{\text{enc}}),
\label{eq4}
\end{equation}
where $f_{\theta}$ represents the encoder function parameterized by $\theta$, and $W_{\text{enc}}$ and $b_{\text{enc}}$ are the encoder’s weight matrix and bias, respectively.
$\sigma$ is a non-linear activation function (\textit{e.g.} ReLU or Sigmoid), and $z \in \mathbb{R}^{d_z}$ is the latent representation where $d_z < d$.

The decoder attempts to reconstruct the original input from $z$ as:
\begin{equation}
\hat{x} = g_{\phi}(z) = \sigma(W_{\text{dec}} z + b_{\text{dec}}),
\label{eq5}
\end{equation}
where $g_{\phi}$ represents the decoder function parameterized by $\phi$, $W_{\text{dec}}$ and $b_{\text{dec}}$ are the decoder’s weight matrix and bias, and $\hat{x}$ is the reconstructed output.

The AutoEncoder is trained by minimizing the reconstruction loss, typically measured using the Mean Squared Error (MSE) as:
\begin{equation}
\mathcal{L}_{\text{AE}} = \| x - \hat{x} \|^2.
\label{eq6}
\end{equation}

\subsection{RMAdapter}
Adapters \cite{Houlsby2019} are small modules that are added to the transformer layers. Instead of tuning the parameters of the whole network, only the adapter parameters and the classifier are trained.
Adapters are structured as bottlenecks with an inner dimension of \( r \ll d \), where \( r \) is referred to as the \textit{rank} of the adapter.
Specifically, the adapter first performs a down-projection from hidden dimension $d$ to dimension \( r \) using weights \( W_{\text{down}} \in \mathbb{R}^{d \times r} \) and biases \( b_{\text{down}} \in \mathbb{R}^{r} \). This is followed by a non-linear activation function \( \sigma \), which is typically a GELU function \cite{Hendrycks2023}, and then an up-projection with weights \( W_{\text{up}} \in \mathbb{R}^{r \times d} \) and biases \( b_{\text{up}} \in \mathbb{R}^{d} \), which maps the representation back to the hidden dimension \( d \) of the transformer layer as:
\begin{equation}
\text{Adapter}(x) = \sigma (x W_{\text{down}} + b_{\text{down}})W_{\text{up}}^{base} + b_{\text{up}}^{base}.
\label{eq7}
\end{equation}

Inspired by the structural similarities between autoencoders and adapters as shown in Equations \ref{eq4}, \ref{eq5}, \ref{eq7}, we propose RMAdapter as a novel integration as shown in Figure \ref{fig3} . RMAdapter employs a dual-branch architecture consisting of:  
(1) an adaptation branch, $\text{RMAdapter}_{\text{base}}$, parameterized by an up-projection matrix $W_{\text{up}}^{\text{base}} \in \mathbb{R}^{r \times d}$ and biases $b_{\text{up}}^{\text{base}} \in \mathbb{R}^{r \times d}$,  
and (2) a reconstruction branch, $\text{RMAdapter}_{\text{rec}}$, parameterized by two up-projection matrices, $W_{\text{up1}}^{\text{rec}} \in \mathbb{R}^{r \times r}$ and $W_{\text{up2}}^{\text{rec}} \in \mathbb{R}^{r \times d}$, along with corresponding biases, $b_{\text{up1}}^{\text{rec}} \in \mathbb{R}^{r \times r}$ and $b_{\text{up2}}^{\text{rec}} \in \mathbb{R}^{r \times d}$, for reconstruction.
Notably, we share $W_{\text{down}}$ and $b_{\text{down}}$ across both branches to further enhance the model’s performance. 

This results in the following base adapter module:
\begin{equation}
x_{down} = \sigma (x W_{\text{down}} + b_{\text{down}}),
\end{equation}
\begin{equation}
\text{RMAdapter}_{base}(x) =x_{down} W_{\text{up}}^{base} + b_{\text{up}}^{base},
\end{equation}
\begin{equation}
\text{RMAdapter}_{\text{rec}}(x)= \sigma (x_{down} W_{\text{up1}}^{rec} + b_{\text{up1}}^{rec})W_{\text{up2}}^{rec} + b_{\text{up2}}^{rec}.
\end{equation}
Following settings in MMA \cite{yang2024mma}, the adapter module is added into a few higher-layers $i \in \{k,k+1,...,K\} $ of both image and text encoders. We calculate the reconstruction loss terms $ \mathcal{L}_{\text{rec}}^{V}$ and $\mathcal{L}_{\text{rec}}^{T}$ from the $k$-th transformer block for visual branch and text branches respectively:
\begin{equation}
    \mathcal{L}_{\text{rec}}^{V} = \sum_{i=k}^{K}\| [c_i, E_i] - \text{RMAdapter}_{\text{rec}}([c_i, E_i]) \|^2,
\end{equation}
\begin{equation}
    \mathcal{L}_{\text{rec}}^{T} = \sum_{i=k}^{K}\| W_i - \text{RMAdapter}_{\text{rec}}(W_i) \|^2.
\end{equation}
We tested L2, L1, and cosine objectives; L2 yielded the most stable and consistent results, and we therefore adopt it for our reconstruction module. Then the total reconstruction loss $\mathcal{L}_{\text{rec}}$ are:
\begin{equation}
    \mathcal{L}_{\text{rec}} = \lambda_{1} \mathcal{L}_{\text{rec}}^{V} + \lambda_{2} \mathcal{L}_{\text{rec}}^{T},
\end{equation}
where $\lambda_{1}$ and $\lambda_{2}$ are loss balancing hyper-parameters.
For the image encoder $\mathcal{V}$ and text encoder $\mathcal{T}$, our adaptation branch $\text{Adapter}_{base}$ from the $k$-th transformer block are formulated as:
\begin{equation}
\begin{aligned}
[c_i, E_i] =& \mathcal{V}_i([c_{i-1}, E_{i-1}]) \\
&+ \alpha \text{RMAdapter}^{v_i}_{base}([c_{i-1}, E_{i-1}]),
\end{aligned}
\end{equation}
\begin{equation}
\begin{aligned}
W_i = \mathcal{T}_i(W_{i-1}) + \alpha \text{RMAdapter}^{t_i}_{base}(W_{i-1}),
\end{aligned}
\end{equation}
where $\alpha$ is a scale factor. After obtaining the final text representation $w^a$ and image representation $x^a$ based on RMAdapter, the supervised loss are calculated as:
\begin{equation}
\mathcal{L}_{ce} = -\log \frac{\exp(\operatorname{sim}(x^a, w_y^a)/\tau)}
{\sum_{k=1}^{C} \exp(\operatorname{sim}(x^a, w_k^a)/\tau)}.
\end{equation}
Besides the supervised loss, we impose a constraint on the adapted visual and text features to ensure their consistency $\mathcal{L}_{con}$ with the original CLIP's pre-trained features as:
\begin{equation}
\mathcal{L}_{con} = \lambda_{3} \sum_{i=1}^{d} |x^a -x| + \lambda_{4} \sum_{i=1}^{d} |w^a -w|,
\end{equation}
where  $\lambda_{3}$ and $\lambda_{4}$ are loss balancing hyper-parameters.
Then the final loss is defined as:
\begin{equation}
\mathcal{L} = \mathcal{L}_{ce}+\mathcal{L}_{con}+\mathcal{L}_{\text{rec}}.
\end{equation}

\begin{table*}[!ht]
\begin{minipage}{\linewidth}
\centering
\begin{subtable}[t]{0.25\textwidth}
\centering
\caption{Average}
\label{tab:avg}
\resizebox{\textwidth}{!}{%
\begin{tabular}{l|ccc}
\hline
\textbf{Methods} & Base & Novel & HM \\
\hline
CLIP & 69.34 & 74.22 & 71.70 \\
CoOp & 82.69 & 63.22 & 71.66 \\
Co-CoOp & 80.47 & 71.69 & 75.83 \\
KgCoOp & 80.73 & 73.60 & 77.00 \\
MaPLe & 82.28 & 75.14 & 78.55 \\
TCP & 84.13 & 75.36 & 79.51 \\
PromptSRC & 84.26 & 76.10 & 79.97 \\
MMA & 83.20 & 76.80 & 79.87 \\
CoPrompt & 84.00 & 77.23 & 80.48 \\
RMAdapter & \textbf{84.52} & \textbf{77.36} & \textbf{80.62} \\
\hline
\end{tabular}}
\end{subtable}
\hfill
\begin{subtable}[t]{0.25\textwidth}
\centering
\caption{ImageNet}
\label{tab:imagenet}
\resizebox{\textwidth}{!}{%
\begin{tabular}{l|ccc}
\hline
\textbf{Methods} & Base & Novel & HM \\
\hline
CLIP & 72.43 & 68.14 & 70.22 \\
CoOp & 76.47 & 67.88 & 71.92 \\
Co-CoOp & 75.98 & 70.43 & 73.10 \\
KgCoOp & 75.83 & 69.96 & 72.78 \\
MaPLe & 76.66 & 70.54 & 73.47 \\
TCP & 77.27 & 69.87 & 73.38 \\
PromptSRC & 77.60 & 70.13 & 73.66 \\
MMA & 77.31 & 71.00 & 74.02 \\
CoPrompt & 77.67 & 71.27 & 74.33 \\
RMAdapter & \textbf{77.87} & \textbf{71.50} & \textbf{74.52} \\
\hline
\end{tabular}}
\end{subtable}
\hfill
\begin{subtable}[t]{0.25\textwidth}
\centering
\caption{Caltech101}
\label{tab:caltech}
\resizebox{\textwidth}{!}{%
\begin{tabular}{l|ccc}
\hline
\textbf{Methods} & Base & Novel & HM \\
\hline
CLIP & 96.84 & 94.00 & 95.40 \\
CoOp & 98.00 & 89.81 & 93.73 \\
Co-CoOp & 97.96 & 93.81 & 95.84 \\
KgCoOp & 97.92 & 93.99 & 95.91 \\
MaPLe & 98.10 & 93.99 & 96.02 \\
TCP & 98.23 & 94.67 & 96.42 \\
PromptSRC & 98.27 & 94.00 & 96.08 \\
MMA & \textbf{98.40} & 94.00 & 96.15 \\
CoPrompt & 98.27 & \textbf{94.90} & \textbf{96.55} \\
RMAdapter & \textbf{98.40} & 94.27 & 96.26 \\
\hline
\end{tabular}}
\end{subtable}

\begin{subtable}[t]{0.25\textwidth}
\centering
\caption{OxfordPets}
\label{tab:oxford}
\resizebox{\textwidth}{!}{%
\begin{tabular}{l|ccc}
\hline
\textbf{Methods} & Base & Novel & HM \\
\hline
CLIP & 91.17 & 97.26 & 94.12 \\
CoOp & 93.67 & 95.29 & 94.47 \\
Co-CoOp & 95.20 & 97.69 & 96.43 \\
KgCoOp & 94.65 & 97.76 & 96.18 \\
MaPLe & 95.43 & 97.76 & 96.58 \\
TCP & 94.67 & 97.20 & 95.92 \\
PromptSRC & 95.37 & \textbf{98.10} & 96.72 \\
MMA & 95.40 & 98.07 & 96.72 \\  
CoPrompt & 95.67 & \textbf{98.10} & 96.87 \\
RMAdapter & \textbf{95.70} & \textbf{98.10} & \textbf{96.89} \\
\hline
\end{tabular}}
\end{subtable}
\hfill
\begin{subtable}[t]{0.25\textwidth}
\centering
\caption{StanfordCars}
\label{tab:cars}
\resizebox{\textwidth}{!}{%
\begin{tabular}{l|ccc}
\hline
\textbf{Methods} & Base & Novel & HM \\
\hline
CLIP & 63.37 & 74.89 & 68.65 \\
CoOp & 78.12 & 60.40 & 68.13 \\
Co-CoOp & 70.49 & 73.29 & 72.01 \\
KgCoOp & 71.76 & 75.04 & 73.36 \\
MaPLe & 72.94 & 74.04 & 73.47 \\
TCP & 80.80 & 74.13 & 77.32 \\
PromptSRC & 76.87 & 75.00 & 75.94 \\
MMA & 78.50 & 73.10 & 75.70 \\  
CoPrompt & 76.97 & 74.40 & 75.66 \\
RMAdapter & \textbf{80.60} & \textbf{75.10} & \textbf{77.75} \\
\hline
\end{tabular}}
\end{subtable}
\hfill
\begin{subtable}[t]{0.25\textwidth}
\centering
\caption{Flowers102}
\label{tab:flowers}
\resizebox{\textwidth}{!}{%
\begin{tabular}{l|ccc}
\hline
\textbf{Methods} & Base & Novel & HM \\
\hline
CLIP & 72.08 & \textbf{77.80} & 74.83 \\
CoOp & 97.60 & 59.67 & 74.06 \\
Co-CoOp & 94.87 & 71.75 & 81.76 \\
KgCoOp & 95.92 & 72.60 & 83.06 \\
MaPLe & 95.92 & 72.46 & 82.05 \\
TCP & 97.73 & 75.57 & 85.23 \\
PromptSRC & 97.27 & 76.00 & 85.71 \\
MMA & 97.77 & 75.93 & 85.48 \\   
CoPrompt & 97.27 & 76.60 & 85.71 \\
RMAdapter & \textbf{98.10} & 77.47 & \textbf{86.52} \\
\hline
\end{tabular}}
\end{subtable}

\begin{subtable}[t]{0.25\textwidth}
    \centering
    \caption{Food101}
    \label{tab:food101}
    \resizebox{\textwidth}{!}{%
    \begin{tabular}{l|ccc}
    \hline
    \textbf{Methods} & Base & Novel & HM \\
    \hline
    CLIP & 90.10 & 91.22 & 90.66 \\
    CoOp & 88.33 & 82.26 & 85.19 \\
    Co-CoOp & 90.11 & 91.29 & 90.99 \\
    KgCoOp & 90.71 & 91.70 & 91.19 \\
    MaPLe & 90.71 & 92.05 & 91.38 \\
    TCP & 90.57 & 91.37 & 90.97 \\
    PromptSRC & 90.67 & 91.53 & 91.10 \\ 
    MMA & 90.13 & 91.30 &  90.71 \\  
    CoPrompt & \textbf{90.73} & \textbf{92.07} & \textbf{91.40} \\
    RMAdapter & 90.53 & 91.67 & 91.10 \\
    \hline
    \end{tabular}}
\end{subtable}
\hfill
\begin{subtable}[t]{0.25\textwidth}
    \centering
    \caption{FGVCAircraft}
    \label{tab:fgvcaircraft}
    \resizebox{\textwidth}{!}{%
    \begin{tabular}{l|ccc}
    \hline
    \textbf{Methods} & Base & Novel & HM \\
    \hline
    CLIP & 27.19 & 36.29 & 31.09 \\
    CoOp & 40.44 & 22.30 & 28.75 \\
    Co-CoOp & 33.41 & 23.71 & 27.45 \\
    KgCoOp & 36.21 & 35.12 & 35.83 \\
    MaPLe & 37.44 & 35.61 & 36.50 \\
    TCP & 41.97 & 34.43 & 37.83 \\
    PromptSRC & 42.73 & 37.87 & \textbf{40.15} \\
    MMA & 40.57 & 36.33 & 38.33 \\  
    CoPrompt & 40.20 & \textbf{39.33} & 39.76 \\
    RMAdapter & \textbf{43.20} & 37.20 & 39.76 \\
    \hline
    \end{tabular}}
\end{subtable}
\hfill
\begin{subtable}[t]{0.25\textwidth}
    \centering
    \caption{SUN397}
    \label{tab:sun397}
    \resizebox{\textwidth}{!}{%
    \begin{tabular}{l|ccc}
    \hline
    \textbf{Methods} & Base & Novel & HM \\
    \hline
    CLIP & 69.36 & 75.35 & 72.23 \\
    CoOp & 80.60 & 65.89 & 72.51 \\
    Co-CoOp & 79.74 & 76.86 & 78.27 \\
    KgCoOp & 80.00 & 77.77 & 78.87 \\
    MaPLe & 81.24 & 78.41 & 79.79 \\
    TCP & 82.63 & 78.20 & 80.35 \\
    PromptSRC & \textbf{82.67} & 78.47 & 81.31 \\ 
    MMA & 82.27 & 78.57 & 80.38 \\   
    CoPrompt & 82.63 & \textbf{80.03} & 81.31 \\
    RMAdapter & \textbf{82.87} & 79.97 & \textbf{81.39} \\
    \hline
    \end{tabular}}
\end{subtable}

\begin{subtable}[t]{0.25\textwidth}
\centering
\caption{DTD}
\label{tab:dtd}
\resizebox{\textwidth}{!}{%
\begin{tabular}{l|ccc}
\hline
\textbf{Methods} & Base & Novel & HM \\
\hline
CLIP & 53.24 & 59.90 & 56.37 \\
CoOp & 79.44 & 41.18 & 54.24 \\
Co-CoOp & 77.01 & 56.00 & 64.85 \\
KgCoOp & 77.55 & 54.99 & 64.35 \\
MaPLe & 80.36 & 59.18 & 68.16 \\
TCP & 82.77 & 58.07 & 68.25 \\
PromptSRC & \textbf{83.37} & 62.97 & 71.75 \\
MMA & 83.20 & 65.63 & 73.38 \\  
CoPrompt & 83.13 & 64.73 & 72.79 \\
RMAdapter & 83.37 & \textbf{66.73} & \textbf{74.09} \\
\hline
\end{tabular}}
\end{subtable}
\hfill
\begin{subtable}[t]{0.25\textwidth}
\centering
\caption{EuroSAT}
\label{tab:eurosat}
\resizebox{\textwidth}{!}{%
\begin{tabular}{l|ccc}
\hline
\textbf{Methods} & Base & Novel & HM \\
\hline
CLIP & 56.48 & 64.05 & 60.03 \\
CoOp & 92.19 & 54.74 & 68.69 \\
Co-CoOp & 87.49 & 60.04 & 71.21 \\
KgCoOp & 85.64 & 64.34 & 73.48 \\
MaPLe & 94.07 & 73.23 & 82.30 \\
TCP & 91.63 & 74.73 & 82.32 \\
PromptSRC & 92.90 & 73.90 & 82.32 \\   
MMA & 85.46 & 82.34 &  83.87 \\  
CoPrompt & \textbf{94.60} & \textbf{78.57} & \textbf{85.84} \\
RMAdapter & 92.17 & 78.33 & 84.69 \\
\hline  
\end{tabular}}
\end{subtable}
\hfill
\begin{subtable}[t]{0.25\textwidth}
\centering
\caption{UCF101}
\label{tab:ucf101}
\resizebox{\textwidth}{!}{%
\begin{tabular}{l|ccc}
\hline
\textbf{Methods} & Base & Novel & HM \\
\hline
CLIP & 70.53 & 77.50 & 73.85 \\
CoOp & 84.69 & 56.05 & 67.46 \\
Co-CoOp & 82.33 & 73.45 & 77.64 \\
KgCoOp & 82.89 & 76.67 & 79.65 \\
MaPLe & 83.00 & 78.66 & 80.77 \\
TCP & \textbf{87.13} & \textbf{80.77} & \textbf{83.83} \\
PromptSRC & 87.10 & 78.80 & 82.74 \\    
ProGrad & 86.23 & 80.03 &  82.20 \\  
CoPrompt & 86.90 & 79.57 & 83.07 \\
RMAdapter & 86.87 & 80.63 & 83.63 \\
\hline
\end{tabular}}
\end{subtable}

\end{minipage}
\caption{Comparison with state-of-the-art methods on different datasets in the Base-to-Novel Generalization setting. ``Base'' and ``Novel'' are the recognition accuracies on base and novel classes respectively. ``HM'' is the harmonic mean of base and new accuracy, demonstrating the trade-off between adaption and generalization.}
\label{basetonovel}
\end{table*}

\section{Experiments}
\subsection{Benchmark setting}
To evaluate the proposed method, we adopt the experimental setup and protocols established in \cite{Zhou2022ConditionalPrompt,Zhou2022LearningPrompt}. We follow the hyperparameter settings outlined in \cite{Khattak2023SelfReg, yang2024mma}. To ensure a fair comparison, we select baselines that are implemented under the same experimental settings. We describe training details and evaluation protocols in the Supplementary Material.

\subsection{Base-To-Novel Generalization}
\begin{table*}[h]
    \centering
    \resizebox{0.8\textwidth}{!}{
    \begin{tabular}{lcccccccccccc}
        \toprule
        & \multicolumn{1}{c}{\textbf{Source}} & \multicolumn{10}{c}{\textbf{Target}} & \\
        \cmidrule(lr){2-2} \cmidrule(lr){3-12}
        \textbf{Methods} & \textbf{ImNet} & \textbf{Caltech} & \textbf{Pets} & \textbf{Cars} & \textbf{Flowers} & \textbf{Food} & \textbf{Aircraft} & \textbf{SUN397} & \textbf{DTD} & \textbf{EuroSAT} & \textbf{UCF} & \textbf{Ave.} \\
        \midrule
        CoOp & \textbf{71.51} & 93.70 & 89.14 & 64.51 & 68.71 & 85.30 & 18.47 & 64.15 & 41.92 & 46.39 & 66.55 & 63.88 \\
        Co-CoOp & 71.02 & 94.43 & 90.14 & 65.32 & 71.88 & 86.06 & 22.94 & 67.36 & 45.73 & 45.37 & 68.21 & 65.74 \\
        MaPLe & 70.72 & 93.53 & 90.49 & 65.57 & 72.23 & 86.20 & 24.74 & 67.01 & 46.49 & 48.06 & 68.69 & 66.30 \\
        PromptSRC & 71.27 & 93.60 & 90.25 & \textbf{65.70} & 70.25 & 86.15 & 23.90 & 67.10 & 46.87 & 45.50 & 68.75 & 65.81 \\
        MMA & 71.00 & 93.80 & 90.30 & 66.13 & 72.07 & 86.12 & \textbf{25.33} & 68.17 & 46.57 & 49.24 & 68.32 & 66.61 \\
        CoPrompt & 70.80 & \textbf{94.50} & 90.73 & 65.67 & 72.30 & \textbf{86.43} & 24.00 & 67.57 & 47.07 & \textbf{51.90} & 69.73 & 67.00 \\
        RMAdapter & 71.37 & 94.10 & \textbf{90.85} &\textbf{ 65.80} & \textbf{72.65} & 86.33 & 24.90 & \textbf{68.36} & \textbf{47.33} & 51.50 & \textbf{69.96} & \textbf{67.56} \\
        \bottomrule
    \end{tabular}
    }
    \caption{Performance of RMAdapter on cross-dataset evaluation and its comparison to existing methods. Here, the model is trained on the ImageNet dataset and evaluated on ten other datasets in a zero-shot setting.}
    \label{tabcde}
\end{table*}


From the experimental results in Table \ref{basetonovel}, we draw several key conclusions regarding the effectiveness of our RMAdapter.  
First, across 11 datasets, RMAdapter consistently achieves the highest average performance across all evaluation metrics, including base class accuracy, novel class accuracy, and their harmonic mean (HM).
These results confirm that an effective dual-branch adapter design can simultaneously enhance both task-specific adaptation and generalization.  

Without utilizing any data augmentation or additional modifications to the prompt mechanism, RMAdapter surpasses CoPrompt with an average HM of 80.62, showing a 0.5 improvement in base class accuracy and a 0.13 gain in novel class accuracy. Unlike CoPrompt, RMAdapter adopts a structurally simpler design, which allows it to enhance base class performance more effectively while maintaining competitive generalization to novel classes.
Furthermore, compared to the representative adapter-based method MMA, the introduction of the reconstruction branch and self-consistency constraints in RMAdapter leads to significant performance gains. Specifically, RMAdapter improves base class accuracy, novel class accuracy, and harmonic mean (HM) by 1.32, 0.56, and 0.75, respectively. These findings further validate the effectiveness of incorporating a reconstruction branch and consistency constraints in adapter design, demonstrating that RMAdapter can achieve superior adaptation without compromising generalization performance.

\subsection{Cross-Dataset Evaluation}
We also conduct experiments under the cross-dataset setting. The model is first trained on 1,000 categories of ImageNet and then evaluated in a zero-shot manner on the other 10 datasets used in previous experiments.  
As shown in Table \ref{tabcde}, RMAdapter achieves performance improvements on 5 out of 10 target datasets. Overall, RMAdapter attains an average accuracy of 67.56, outperforming MMA by 0.95 and CoPrompt by 0.56.  
Additionally, RMAdapter demonstrates highly competitive performance on the source dataset, ImageNet, surpassing MMA by 0.37 and CoPrompt by 0.57. These results further validate the superior zero-shot transferability of RMAdapter, confirming its effectiveness in adapting to unseen datasets.

\subsection{Domain Generalization}
In Table \ref{tab:dg}, we present the domain generalization experimental results. In this setting, the original ImageNet dataset is used as the source dataset for fine-tuning, and the model is then evaluated on four ImageNet variants that originate from different distributions.  
In this evaluation, our RMAdapter achieves the best performance on three out-of-domain datasets, outperforming Bayesian Prompt by 0.23 and CoPrompt by 0.29. These results demonstrate the robustness of RMAdapter in handling domain shifts, further validating its effectiveness in domain generalization task.

\begin{table}[h]
    \centering
    \resizebox{0.45\textwidth}{!}{ 
    \begin{tabular}{lccccccc}
        \toprule
        & \multicolumn{1}{c}{\textbf{Source}} & \multicolumn{5}{c}{\textbf{Target}} & \\
        \cmidrule(lr){2-2} \cmidrule(lr){3-7}
        \textbf{Method} & \textbf{ImNet} & \textbf{-V2} & \textbf{-S} & \textbf{-A} & \textbf{-R} & \textbf{Ave.} \\
        \midrule
        CLIP & 66.73 & 60.83 & 46.15 & 47.77 & 73.96 & 57.17 \\
        CoOp & 71.51 & 64.20 & 47.99 & 49.71 & 75.21 & 59.28 \\
        Co-CoOp & 71.02 & 64.07 & 48.75 & 50.63 & 76.18 & 59.90 \\
        KgCoOp & 71.20 & 64.10 & 48.97 & 50.69 & 76.70 & 60.11 \\
        MaPLe & 70.72 & 64.07 & 49.15 & 50.90 & 76.98 & 60.26 \\
        PromptSRC & 71.27 & 64.35 & \textbf{49.55} & 50.90 & \textbf{77.80} & 60.65 \\     
         MMA & 71.00 & 64.33 & 49.13 & 51.12 & 77.32 & 60.48  \\
        CoPrompt & 70.80 & 64.25 & 49.43 & 50.50 & 77.51 & 60.42 \\
        RMAdapter & 71.37 & \textbf{64.45} & 49.50 & \textbf{51.21} & 77.70 & \textbf{60.71} \\
        \bottomrule
    \end{tabular}}
    \caption{Performance on domain generalization.}
    \label{tab:dg}
\end{table}

\subsection{Ablation Study}
\textbf{Effectiveness of reconstruction adapter branch.}
We evaluate the effectiveness of different adapter design choices. These design variants include adding the reconstruction adapter branch to either the vision (V) or language (L) branch individually, as well as incorporating self-consistency constraints. This evaluation aims to assess the impact of the reconstruction adapter branch and self-consistency constraints on adapter performance.  
In Table \ref{tabab1}, we present the average results across 11 classification datasets. The experimental findings indicate that the reconstruction adapter branch benefits both branches. Furthermore, when self-consistency constraints are applied alongside the reconstruction adapter branch, the model achieves improved performance on base classes without compromising the accuracy of novel classes.
In this setting, the self-consistency constraints  provide global alignment, while the reconstruction branch enables fine-grained preservation.
\begin{table}[h]
    \centering
    \resizebox{0.45\textwidth}{!}{
    \begin{tabular}{lccc}
        \toprule
        \textbf{Models} & \textbf{Base} & \textbf{Novel} & \textbf{HM} \\
        \midrule
        No Reconstruction and Constraints (MMA) & 83.20 & 76.80 & 79.87 \\
        MMA + Constraints & 83.86 & 76.31 & 79.91  \\
        MMA + Constraints + Text Reconstruction & 84.13 & 77.18 & 80.51  \\
        MMA + Constraints + Visual Reconstruction & 84.20 & 76.91 & 80.39 \\
        RMAdapter & \textbf{84.52} & \textbf{77.36} & \textbf{80.62} \\
        \bottomrule
    \end{tabular}}
    \caption{Performance with different model variants on the reconstruction adapter branch.}
    \label{tabab1}
\end{table}
\\
\textbf{Design strategies for reconstruction branch.}
The reconstruction branch is designed to preserve the model’s overall knowledge structure and generalization ability by constraining the modifications introduced by the adapter while adapting to new tasks.
To investigate the optimal design, we compare three integration strategies: (1) sharing the down-projection layer, (2) sharing the up-projection layer, and (3) using independent down- and up-projection layers, noting that effective regularization only occurs when the projection is shared. The results, summarized in Table \ref{tababshare}, reveal that sharing the down-projection layer yields the best performance.
In contrast, sharing the up-projection layer compromises task adaptation, as the up-projection struggles to simultaneously reconstruct lost information and adapt effectively to downstream tasks. Meanwhile, the independent down- and up-projection structure fails to achieve satisfactory results, likely due to the limited sample size in few-shot settings.
\begin{table}[h]
    \centering
    \resizebox{0.45\textwidth}{!}{
    \begin{tabular}{lccc}
        \toprule
        \textbf{Models} & \textbf{Base} & \textbf{Novel} & \textbf{HM} \\
        \midrule
        Independent  & 84.25 & 77.11 & 80.52 \\
        Shared up-projection & 83.96 & 76.95 & 80.30  \\
        Shared Down-projection (RMAdapter)  & \textbf{84.52} & \textbf{77.36} & \textbf{80.62} \\
        \bottomrule
    \end{tabular}}
    \caption{Performance with different design strategies for integrating the reconstruction branch.}
    \label{tababshare}
\end{table}

\textbf{Number of layers for reconstruction branch.}
We further investigate the impact of the number of linear layers in the up-projection layer. 
The results, presented in Table \ref{tabnumlayer}, indicate that the two-layer adapter achieves slightly better performance than the single-layer design. This suggests that adding an additional layer allows the model to capture more complex relationships and enhance reconstruction accuracy to some extent. However, employing a three-layer adapter leads to a significant drop in performance. We attribute this to the increased number of parameters, which may reintroduce the overfitting problem in few-shot learning scenarios due to the limited availability of training samples.


\begin{table}[h]
    \centering
    \resizebox{0.3\textwidth}{!}{
    \begin{tabular}{lccc}
        \toprule
        \textbf{Num of Layers } & \textbf{Base} & \textbf{Novel} & \textbf{HM} \\
        \midrule
        1 & 84.36 & 77.11 & 80.57 \\
        2 (RMAdapter) & \textbf{84.52} & \textbf{77.36} & \textbf{80.62} \\
        3  & 83.65 & 76.81 & 80.08 \\
        \bottomrule
    \end{tabular}}
    \caption{Performance with different number of layers.}
    \label{tabnumlayer}
\end{table}

\textbf{Overhead of reconstruction branch.}
RMAdapter is designed as a lightweight and parameter-efficient adaptation method. The reconstruction branch, added to preserve generalizable features, introduces only minimal additional overhead. Table~\ref{tab:efficiency_summary} highlights both the quantitative cost of this branch and the architectural strategies that support overall efficiency.
\begin{table}[h]
    \centering
    \footnotesize
    \resizebox{\linewidth}{!}{
    \begin{tabular}{l|c||l|l}
        \toprule
        \multicolumn{2}{c||}{\textbf{(a) Quantitative Overhead}} 
        & \multicolumn{2}{c}{\textbf{(b) Efficiency-Oriented Design}} \\
        \midrule
        \textbf{Metric} & \textbf{Value} 
        & \textbf{Strategy} & \textbf{Effect} \\
        \midrule
        Params (added)   & 320K   & Shared Down-projection   & Fewer params \\
        GPU Memory       & +3\%   & Local Reconstruction Loss    & Low cost \\
        Train Time       & +5\%   & Layer-wise Insertion  & Efficient adaptation \\

        \bottomrule
    \end{tabular}
    }
    \caption{Efficiency summary: (a) overhead from the reconstruction branch; (b) design strategies enabling low-cost adaptation.}
    \label{tab:efficiency_summary}
\end{table}

\section{Conclusion}
In this paper, we propose the RMAdapter, a reconstruction-based multimodal adapter for adapting pre-trained VLMs in few-shot scenarios. Its dual-branch design balances task-specific adaptation and universal knowledge retention, ensuring strong zero-shot generalization while improving few-shot learning.
RMAdapter is carefully designed to remain lightweight, with minimal computational overhead due to local loss computation and parameter sharing.
Experiments on three tasks demonstrate that RMAdapter consistently outperforms state-of-the-art methods. Future work will explore scalability, integration with prompt-based tuning.

\section{Acknowledgments}
This work was supported in part by the National Key R\&D Program of China (2022ZD0161901), the National Natural Science Foundation of China  (82441024), the Beijing Nova Program (20230484297), the Research Program of State Key Laboratory of Complex and Critical Software Environment, and the Fundamental Research Funds for the Central Universities.

\bibliography{aaai2026}

\begin{thebibliography}{34}
\providecommand{\natexlab}[1]{#1}

\bibitem[{Bahng et~al.(2022)Bahng, Jahanian, Sankaranarayanan, and Isola}]{Bahng2022}
Bahng, H.; Jahanian, A.; Sankaranarayanan, S.; and Isola, P. 2022.
\newblock Visual prompting: Modifying pixel space to adapt pre-trained models.
\newblock \emph{arXiv preprint}, arXiv:2203.17274.

\bibitem[{Baldi(2012)}]{pmlr-v27-baldi12a}
Baldi, P. 2012.
\newblock Autoencoders, Unsupervised Learning, and Deep Architectures.
\newblock In Guyon, I.; Dror, G.; Lemaire, V.; Taylor, G.; and Silver, D., eds., \emph{Proceedings of ICML Workshop on Unsupervised and Transfer Learning}, volume~27 of \emph{Proceedings of Machine Learning Research}, 37--49. Bellevue, Washington, USA: PMLR.

\bibitem[{Bulat and Tzimiropoulos(2023)}]{Bulat2023}
Bulat, A.; and Tzimiropoulos, G. 2023.
\newblock Lasp: Text-to-text optimization for language-aware soft prompting of vision language models.
\newblock In \emph{Proceedings of the IEEE/CVF Conference on Computer Vision and Pattern Recognition}, 23232--23241.

\bibitem[{Chen et~al.(2022{\natexlab{a}})Chen, Yang, Lai, and Xie}]{Chen2022}
Chen, Q.; Yang, L.; Lai, J.-H.; and Xie, X. 2022{\natexlab{a}}.
\newblock Self-supervised image-specific prototype exploration for weakly supervised semantic segmentation.
\newblock In \emph{Proceedings of the IEEE/CVF Conference on Computer Vision and Pattern Recognition}, 4288--4298.

\bibitem[{Chen et~al.(2022{\natexlab{b}})Chen, Ge, Tong, Wang, Song, Wang, and Luo}]{Chen2022Adaptformer}
Chen, S.; Ge, C.; Tong, Z.; Wang, J.; Song, Y.; Wang, J.; and Luo, P. 2022{\natexlab{b}}.
\newblock Adaptformer: Adapting vision transformers for scalable visual recognition.
\newblock \emph{Advances in Neural Information Processing Systems}, 35: 16664--16678.

\bibitem[{Farina et~al.(2025)Farina, Mancini, Iacca, and Ricci}]{farina2025rethinking}
Farina, M.; Mancini, M.; Iacca, G.; and Ricci, E. 2025.
\newblock Rethinking Few-Shot Adaptation of Vision-Language Models in Two Stages.
\newblock In \emph{Proceedings of the Computer Vision and Pattern Recognition Conference}, 29989--29998.

\bibitem[{Gao et~al.(2023)Gao, Geng, Zhang, Ma, Fang, Zhang, Li, and Qiao}]{Gao2023}
Gao, P.; Geng, S.; Zhang, R.; Ma, T.; Fang, R.; Zhang, Y.; Li, H.; and Qiao, Y. 2023.
\newblock Clip-adapter: Better vision-language models with feature adapters.
\newblock \emph{International Journal of Computer Vision}, 1--15.

\bibitem[{Gu et~al.(2021)Gu, Lin, Kuo, and Cui}]{Gu2021}
Gu, X.; Lin, T.-Y.; Kuo, W.; and Cui, Y. 2021.
\newblock Open-vocabulary object detection via vision and language knowledge distillation.
\newblock \emph{arXiv preprint}, arXiv:2104.13921.

\bibitem[{Guo and Gu(2025)}]{guo2025mmrl}
Guo, Y.; and Gu, X. 2025.
\newblock Mmrl: Multi-modal representation learning for vision-language models.
\newblock In \emph{Proceedings of the Computer Vision and Pattern Recognition Conference}, 25015--25025.

\bibitem[{Hendrycks and Gimpel(2023)}]{Hendrycks2023}
Hendrycks, D.; and Gimpel, K. 2023.
\newblock Gaussian error linear units (GELUs).
\newblock \emph{arXiv preprint}, arXiv:1606.08415.

\bibitem[{Houlsby et~al.(2019)Houlsby, Giurgiu, Jastrzebski, Morrone, Laroussilhe, Gesmundo, Attariyan, and Gelly}]{Houlsby2019}
Houlsby, N.; Giurgiu, A.; Jastrzebski, S.; Morrone, B.; Laroussilhe, Q.~D.; Gesmundo, A.; Attariyan, M.; and Gelly, S. 2019.
\newblock Parameter-efficient transfer learning for NLP.
\newblock In \emph{International Conference on Machine Learning}, 2790--2799. PMLR.

\bibitem[{Hu et~al.(2021)Hu, Wallis, Allen-Zhu, Li, Wang, Wang, and et~al.}]{Hu2021}
Hu, E.~J.; Wallis, P.; Allen-Zhu, Z.; Li, Y.; Wang, S.; Wang, L.; and et~al., W.~C. 2021.
\newblock LoRA: Low-rank adaptation of large language models.
\newblock In \emph{International Conference on Learning Representations}.

\bibitem[{Jia et~al.(2021)Jia, Yang, Xia, Chen, Parekh, Pham, Le, Sung, Li, and Duerig}]{Jia2021}
Jia, C.; Yang, Y.; Xia, Y.; Chen, Y.-T.; Parekh, Z.; Pham, H.; Le, Q.; Sung, Y.-H.; Li, Z.; and Duerig, T. 2021.
\newblock Scaling up visual and vision-language representation learning with noisy text supervision.
\newblock In \emph{International Conference on Machine Learning}, 4904--4916. PMLR.

\bibitem[{Jia et~al.(2022)Jia, Tang, Chen, Cardie, Belongie, Hariharan, and Lim}]{Jia2022}
Jia, M.; Tang, L.; Chen, B.-C.; Cardie, C.; Belongie, S.; Hariharan, B.; and Lim, S.-N. 2022.
\newblock Visual prompt tuning.
\newblock In \emph{European Conference on Computer Vision}.

\bibitem[{Jiang et~al.(2022)Jiang, Zhang, Huang, Ge, Ni, Lu, Zhou, Song, and Huang}]{Jiang2022}
Jiang, H.; Zhang, J.; Huang, R.; Ge, C.; Ni, Z.; Lu, J.; Zhou, J.; Song, S.; and Huang, G. 2022.
\newblock Cross-modal adapter for text-video retrieval.
\newblock \emph{arXiv preprint}, arXiv:2211.09623.

\bibitem[{Khattak et~al.(2023{\natexlab{a}})Khattak, Rasheed, Maaz, Khan, and Khan}]{Khattak2023}
Khattak, M.~U.; Rasheed, H.; Maaz, M.; Khan, S.; and Khan, F.~S. 2023{\natexlab{a}}.
\newblock Maple: Multi-modal prompt learning.
\newblock In \emph{Proceedings of the IEEE/CVF Conference on Computer Vision and Pattern Recognition}, 19113--19122.

\bibitem[{Khattak et~al.(2023{\natexlab{b}})Khattak, Wasim, Naseer, Khan, Yang, and Khan}]{Khattak2023SelfReg}
Khattak, M.~U.; Wasim, S.~T.; Naseer, M.; Khan, S.; Yang, M.-H.; and Khan, F.~S. 2023{\natexlab{b}}.
\newblock Self-regulating prompts: Foundational model adaptation without forgetting.
\newblock In \emph{Proceedings of the IEEE/CVF International Conference on Computer Vision}, 15190--15200.

\bibitem[{Lester, Al-Rfou, and Constant(2021)}]{Lester2021}
Lester, B.; Al-Rfou, R.; and Constant, N. 2021.
\newblock The power of scale for parameter-efficient prompt tuning.
\newblock In \emph{Proceedings of the Conference on Empirical Methods in Natural Language Processing}, 3045--3059.

\bibitem[{Mou et~al.(2023)Mou, Wang, Xie, Zhang, Qi, Shan, and Qie}]{Mou2023}
Mou, C.; Wang, X.; Xie, L.; Zhang, J.; Qi, Z.; Shan, Y.; and Qie, X. 2023.
\newblock T2i-adapter: Learning adapters to dig out more controllable ability for text-to-image diffusion models.
\newblock \emph{arXiv preprint}, arXiv:2302.08453.

\bibitem[{Oord, Li, and Vinyals(2018)}]{Oord2018}
Oord, A. V.~D.; Li, Y.; and Vinyals, O. 2018.
\newblock Representation learning with contrastive predictive coding.
\newblock \emph{arXiv preprint}, arXiv:1807.03748.

\bibitem[{Radford et~al.(2021)Radford, Kim, Hallacy, Ramesh, Goh, Agarwal, Sastry, Askell, Mishkin, Clark et~al.}]{radford2021learning}
Radford, A.; Kim, J.~W.; Hallacy, C.; Ramesh, A.; Goh, G.; Agarwal, S.; Sastry, G.; Askell, A.; Mishkin, P.; Clark, J.; et~al. 2021.
\newblock Learning transferable visual models from natural language supervision.
\newblock In \emph{International conference on machine learning}, 8748--8763. PmLR.

\bibitem[{Roy and Etemad(2024)}]{roy2024consistencyguidedpromptlearningvisionlanguage}
Roy, S.; and Etemad, A. 2024.
\newblock Consistency-guided Prompt Learning for Vision-Language Models.
\newblock arXiv:2306.01195.

\bibitem[{Tsimpoukelli et~al.(2021)Tsimpoukelli, Menick, Cabi, Eslami, Vinyals, and Hill}]{Tsimpoukelli2021}
Tsimpoukelli, M.; Menick, J.~L.; Cabi, S.; Eslami, S.; Vinyals, O.; and Hill, F. 2021.
\newblock Multimodal few-shot learning with frozen language models.
\newblock \emph{Advances in Neural Information Processing Systems}, 34: 200--212.

\bibitem[{Yang et~al.(2024)Yang, Zhang, Wang, and Xie}]{yang2024mma}
Yang, L.; Zhang, R.-Y.; Wang, Y.; and Xie, X. 2024.
\newblock Mma: Multi-modal adapter for vision-language models.
\newblock In \emph{Proceedings of the IEEE/CVF Conference on Computer Vision and Pattern Recognition}, 23826--23837.

\bibitem[{Yao, Zhang, and Xu(2023)}]{yao2023visual}
Yao, H.; Zhang, R.; and Xu, C. 2023.
\newblock Visual-language prompt tuning with knowledge-guided context optimization.
\newblock In \emph{Proceedings of the IEEE/CVF conference on computer vision and pattern recognition}, 6757--6767.

\bibitem[{Yao, Zhang, and Xu(2024)}]{yao2024tcp}
Yao, H.; Zhang, R.; and Xu, C. 2024.
\newblock Tcp: Textual-based class-aware prompt tuning for visual-language model.
\newblock In \emph{Proceedings of the IEEE/CVF Conference on Computer Vision and Pattern Recognition}, 23438--23448.

\bibitem[{Yao et~al.(2021)Yao, Huang, Hou, Lu, Niu, Xu, Liang, Li, Jiang, and Xu}]{Yao2021Filip}
Yao, L.; Huang, R.; Hou, L.; Lu, G.; Niu, M.; Xu, H.; Liang, X.; Li, Z.; Jiang, X.; and Xu, C. 2021.
\newblock Filip: Fine-grained interactive language-image pre-training.
\newblock In \emph{International Conference on Learning Representations}.

\bibitem[{Yu et~al.(2022)Yu, Wang, Vasudevan, Yeung, Seyedhosseini, and Wu}]{Yu2022}
Yu, J.; Wang, Z.; Vasudevan, V.; Yeung, L.; Seyedhosseini, M.; and Wu, Y. 2022.
\newblock Coca: Contrastive captioners are image-text foundation models.
\newblock \emph{arXiv preprint}, arXiv:2205.01917.

\bibitem[{Yuan et~al.(2021)Yuan, Chen, Chen, Codella, Dai, Gao, Hu, Huang, Li, and et~al.}]{Yuan2021}
Yuan, L.; Chen, D.; Chen, Y.-L.; Codella, N.; Dai, X.; Gao, J.; Hu, H.; Huang, X.; Li, B.; and et~al., C.~L. 2021.
\newblock Florence: A new foundation model for computer vision.
\newblock \emph{arXiv preprint}, arXiv:2111.11432.

\bibitem[{Zhai et~al.(2022)Zhai, Wang, Mustafa, Steiner, Keysers, Kolesnikov, and Beyer}]{Zhai2022}
Zhai, X.; Wang, X.; Mustafa, B.; Steiner, A.; Keysers, D.; Kolesnikov, A.; and Beyer, L. 2022.
\newblock Lit: Zero-shot transfer with locked-image text tuning.
\newblock In \emph{Proceedings of the IEEE/CVF Conference on Computer Vision and Pattern Recognition}, 18123--18133.

\bibitem[{Zhang et~al.(2021)Zhang, Fang, Zhang, Gao, Li, Dai, Qiao, and Li}]{Zhang2021TipAdapter}
Zhang, R.; Fang, R.; Zhang, W.; Gao, P.; Li, K.; Dai, J.; Qiao, Y.; and Li, H. 2021.
\newblock Tip-adapter: Training-free clip-adapter for better vision-language modeling.
\newblock \emph{arXiv preprint}, arXiv:2111.03930.

\bibitem[{Zhou et~al.(2022{\natexlab{a}})Zhou, Yang, Loy, and Liu}]{Zhou2022ConditionalPrompt}
Zhou, K.; Yang, J.; Loy, C.~C.; and Liu, Z. 2022{\natexlab{a}}.
\newblock Conditional prompt learning for vision-language models.
\newblock In \emph{Proceedings of the IEEE/CVF Conference on Computer Vision and Pattern Recognition}, 16816--16825.

\bibitem[{Zhou et~al.(2022{\natexlab{b}})Zhou, Yang, Loy, and Liu}]{Zhou2022LearningPrompt}
Zhou, K.; Yang, J.; Loy, C.~C.; and Liu, Z. 2022{\natexlab{b}}.
\newblock Learning to prompt for vision-language models.
\newblock \emph{International Journal of Computer Vision}, 130(9): 2337--2348.

\bibitem[{Zhou et~al.(2022{\natexlab{c}})Zhou, Girdhar, Joulin, Krähénbühl, and Misra}]{Zhou2022Detecting}
Zhou, X.; Girdhar, R.; Joulin, A.; Krähénbühl, P.; and Misra, I. 2022{\natexlab{c}}.
\newblock Detecting twenty-thousand classes using image-level supervision.
\newblock In \emph{European Conference on Computer Vision}, 350--368. Springer.

\end{thebibliography}

\end{document}